%% file: main.tex
\documentclass[conference]{IEEEtran}
\usepackage{times}

\usepackage[numbers,sort&compress]{natbib}
\usepackage{multicol}
\usepackage{multirow}
\usepackage[bookmarks=true]{hyperref}
\usepackage{xspace}
\usepackage{xcolor}
\usepackage{graphicx}
\usepackage{caption}
\usepackage{subcaption}
\usepackage{adjustbox}
\usepackage{xfrac}
\usepackage{listings}
\usepackage{float}
\usepackage{booktabs}
\usepackage{bm}
\usepackage{amsfonts}

\newcommand{\sysName}{\textsc{DexCap}\xspace}
\newcommand{\algoName}{\textsc{DexIL}\xspace}
\newcommand{\q}[1]{$\mathbf{\mathcal{Q}#1}$}

\usepackage{soul}
\soulregister\citet7
\soulregister\citep7

\begin{document}

\title{DexCap: Scalable and Portable Mocap Data Collection System for Dexterous Manipulation}

\author{Chen Wang, Haochen Shi, Weizhuo Wang, Ruohan Zhang, Li Fei-Fei, C. Karen Liu\\
Stanford University\\
\authorblockA{\textbf{\textcolor{magenta}{\url{https://dex-cap.github.io}}}}
}

\input{figs/pull}

\begin{abstract}
Imitation learning from human hand motion data presents a promising avenue for imbuing robots with human-like dexterity in real-world manipulation tasks. Despite this potential, substantial challenges persist, particularly with the portability of existing hand motion capture (mocap) systems and the complexity of translating mocap data into effective robotic policies. To tackle these issues, we introduce \sysName, a portable hand motion capture system, alongside \algoName, a novel imitation algorithm for training dexterous robot skills directly from human hand mocap data. \sysName offers precise, occlusion-resistant tracking of wrist and finger motions based on SLAM and electromagnetic field together with 3D observations of the environment. Utilizing this rich dataset, \algoName employs inverse kinematics and point cloud-based imitation learning to seamlessly replicate human actions with robot hands. Beyond direct learning from human motion, \sysName also offers an optional human-in-the-loop correction mechanism during policy rollouts to refine and further improve task performance. Through extensive evaluation across six challenging dexterous manipulation tasks, our approach not only demonstrates superior performance but also showcases the system's capability to effectively learn from in-the-wild mocap data, paving the way for future data collection methods in the pursuit of human-level robot dexterity.\end{abstract}

\IEEEpeerreviewmaketitle

\section{Introduction}
\label{sec:intro}

Building robotic systems to perform everyday manipulation tasks is a long-standing challenge. Our living environments and daily objects are designed with human hand functionality in mind, posing a substantial challenge for developing future home robots. Recent breakthroughs in robotic dexterity, especially in the control of multi-fingered mechanical hands with a high degree of freedom, have shown remarkable potential~\cite{handa2020dexpilot, chen2022visual, videodex}. However, enabling robotic hands to emulate human-level dexterity in manipulation tasks remains unsolved, due to both hardware and algorithmic challenges.

Imitation Learning (IL) \cite{schaal1999imitation,hussein2017imitation} has recently made considerable strides toward this goal~\cite{guzey2023dexterity, sivakumar2022robotic}, especially through supervised training using human demonstration data. One commonly used way to collect data is to teleoperate robot hands to perform the tasks. However, due to the requirement of a real robot system and slow robot motion, this approach is expensive to scale up. An alternative way is to directly track human hand motions during manipulation without controlling the robot. Current system is primarily vision-based with a single-view camera. However, besides the question of whether the tracking algorithm can provide accurate 3D information which is critical for robot policy learning, these systems are vulnerable to visual occlusions that frequently occur during hand-object interactions. 

A better alternative to vision-based methods for gathering dexterous manipulation data is through motion capture (mocap). Mocap systems provides accurate 3D information and are robust to visual occlusions. Hence human operators can directly interact with the environment with their hands, which is fast and easier to scale up since no robot hardware is required. To scale up hand mocap systems to data collection in everyday tasks and environments for robot learning, a suitable system should ideally be portable and robust for long capture sessions, provide accurate finger and wrist poses, as well as 3D environment information. Most hand mocap systems are not portable and rely on well-calibrated third-view cameras. While electromagnetic field (EMF) gloves overcome this issue, they cannot track the 6-DoF wrist pose in the world frame, which is important for end-effectors policy learning. Devices like IMU-based whole-body suits can monitor wrist position but are prone to drift over time.

In addition to hardware challenges, there are also algorithmic challenges to use motion capture data for robot imitation learning.
While dexterous robot hands enable the possibility of learning directly from human hand data, the inherent differences in size, proportion, and kinematic structure between the robot hand and human hand call for innovative algorithms to overcome these embodiment gaps. Towards solving these challenges, our work simultaneously introduces a new portable hand mocap system, \sysName, and an imitation algorithm, \algoName, that allows the robot to learn dexterous manipulation policies directly from the human hand mocap data.

\sysName (Fig.~\ref{fig:pull}) is a portable hand mocap system that tracks the 6-DoF poses of the wrist and the finger motions in real-time (60Hz). The system includes a mocap glove to track finger joints, a camera mounted on top of each glove to track the 6-DoF poses of the wrists with SLAM, and an RGB-D LiDAR camera on the chest to observe the 3D environments.

Besides the hardware challenges, research efforts on developing algorithms to utilize mocap data for robot learning have been missing due to the lack of such a data collection system and collected data. Prior algorithms that learn from human motion focus on learning the rewards~\cite{smith2019avid, bahl2022human}, high-level plans~\cite{wang2023mimicplay, bharadhwaj2023towards}, and visual representations~\cite{nair2022r3m, xiao2022masked}, which often require additional robot data and cannot be directly used for low-level control. In this work, we argue that the main challenge of learning low-level control from human motion is that the data is missing precise 3D information of the hand motion (e.g., 6-DoF hand pose, 3D finger positioning), which are exactly what \sysName can provide.

To leverage data collected by \sysName for learning dexterous robot policies, we propose imitation learning from mocap data, \algoName, which consists of two major steps --- data retargeting and training generative-based behavior cloning policy with point cloud inputs, with an optional human-in-the-loop motion correction step. For retargeting, we use inverse kinematics (IK) to retarget the robotic hand's fingertips to the same 3D location as the human's fingertips. The 6-DoF pose of the wrist is used to initialize the IK to ensure the same wrist motion between the human and the robots. Then we convert RGB-D observations to point cloud-based representations. We then use a point cloud-based behavior cloning algorithm based on Diffusion Policy~\cite{chi2023diffusion}. In more challenging tasks when IK is insufficient to fulfill the embodiment gap between human and robot hands, we propose a human-in-the-loop motion correction mechanism. During policy rollouts, humans can wear the \sysName and interrupt the robot's motion when unexpected behavior occurs, and such interruption data can be further used for policy finetuning. 

In summary, the main contributions of this work include:
\begin{itemize}
    \item \sysName: a novel portable human hand mocap system, enabling real-time tracking of wrist and finger movements for dexterous manipulation tasks.
    \item \algoName: an imitation learning framework leveraging hand mocap data for directly learning dexterous manipulation skills from human hand motions.
    \item Human-in-the-Loop Correction: a human-in-the-loop correction mechanism with \sysName, significantly enhancing robot performance in complex tasks.
\end{itemize}

\section{Related Works}
\label{sec:related}

\begin{figure*}[t]
\includegraphics[width=2.0\columnwidth]{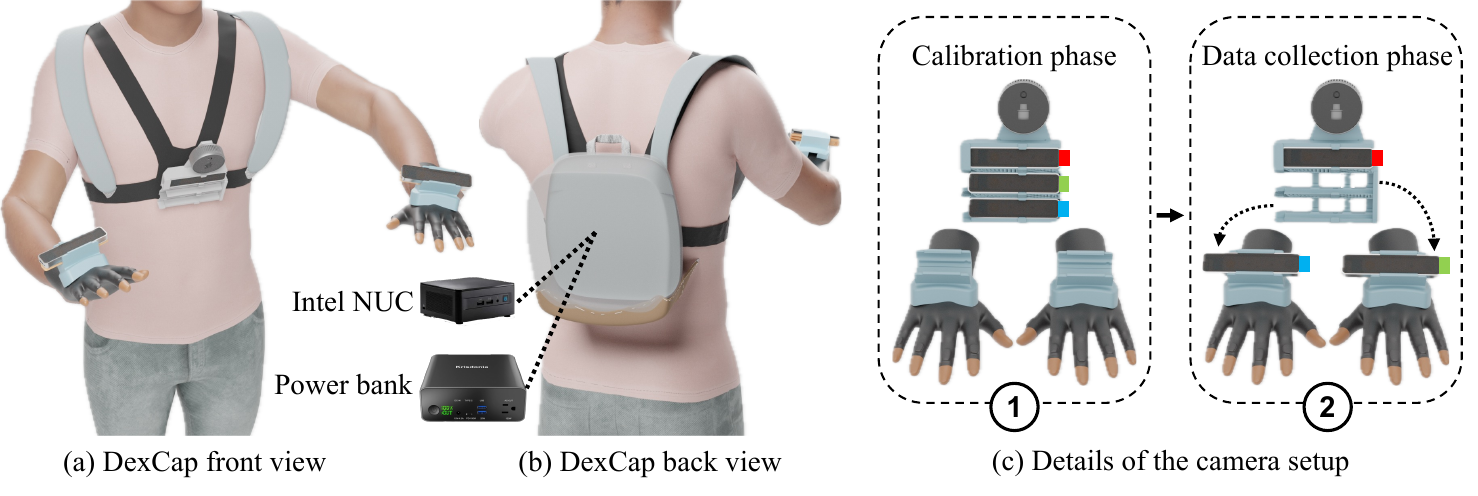}
\vspace{-0.2cm}
\caption{\textbf{Details of the human system.} (a) Our setup includes a 3D-printed rack on a chest harness, featuring a Realsense L515 LiDAR camera on top and three Realsense T265 tracking cameras below. (b) An Intel NUC and power bank in a backpack power the system for approximately 40 minutes of data collection. (c) The T265 cameras, initially in a known pose for calibration, are relocated to hand mounts during data collection to monitor palm positions, ensuring consistency through a click-in design. Finger motions are captured by Rokoko gloves, accurately tracking the finger joint positions.}
\vspace{-0.5cm}
\label{fig:sys}
\end{figure*}

\subsection{Dexterous manipulation}
Dexterous manipulation has been a long-standing research area in robotics~\cite{salisbury1982articulated, mason1985robot, mordatch2012contact, bai2014dexterous, kumar2014real}, posing significant challenges to planning and control due to the high degrees-of-freedom. The traditional optimal control methods~\cite{mordatch2012contact, bai2014dexterous, kumar2014real} often necessitate simplification of the contacts, which is usually not tenable in more complex tasks. Recently, reinforcement learning has been explored to learn dexterous policies in simulation with minimal assumptions about the task or the environment~\cite{handa2022dextreme, chen2022visual, chen2021system, Abhishek2021resetfree, yin2023rotating, qi2022hand, Khandate-RSS-23, huang2023dynamic, chen2023sequential, pitz2023dextrous, xu2023dexterous}. The learned policies can solve complex tasks, including in-hand object re-orientatation~\cite{handa2022dextreme, chen2022visual, yin2023rotating, qi2022hand, Khandate-RSS-23, pitz2023dextrous}, bimanual manipulation~\cite{huang2023dynamic, lin2024twisting}, and long-horizon manipulation~\cite{Abhishek2021resetfree, chen2023sequential}. However, due to the sim-to-real gap, deploying the learned policy on a real-world robot remains challenging. Imitation learning, on the other hand, focuses on learning directly from real-world demonstration data, which is obtained through either teleportation~\cite{handa2020dexpilot, arunachalam2023dexterous, arunachalam2023holo, guzey2023dexterity} or human videos~\cite{qin2022dexmv, mandikal2022dexvip, videodex}. DIME~\cite{arunachalam2023dexterous} uses VR to teleoperate a dexterous hand for data collection; ~\citet{qin2022one} uses an RGB camera to track hand pose for teleoperation; DexTransfer~\cite{chen2022dextransfer} uses human mocap data to guide dexterous grasping; DexMV~\cite{qin2022dexmv}, DexVIP~\cite{mandikal2022dexvip} and VideoDex~\cite{videodex} leverages human video data for learning the motion priors but often require additional training in simulation or real robot teleoperation data. Our work focuses on dexterous imitation learning, which relies on \sysName to collect high-quality hand mocap data grounded in 3D point cloud observation, which can be directly used to train low-level positional control on robots with single or dual hands.

\subsection{Hand motion capture system}
Human hand mocap is an important technique for applications in computer vision and graphics. Most previous systems are camera-based, IMU-based, or electromagnet(EMF)-based. Camera-based systems utilize monocular camera~\cite{zimmermann2019freihand, moon2020interhand2, pavlakos2023reconstructing}, RGB-D camera~\cite{schmidt2014dart, hampali2020honnotate, chao2021dexycb}, VR headset~\cite{han2020megatrack}, or multi-view camera with markers~\cite{taheri2020grab, fan2023arctic}. However, the quality of hand motion tracking quickly deteriorates in scenarios involving heavy occlusions, which happen frequently in hand-object interactions. Some of these systems also require third-view calibrated cameras which are not portable or scalable. More recently, Inertia Measurement Unit (IMU) has been used for in-the-wild human mocap~\cite{huang2018deep, jiang2022transformer, yi2022physical, van2023diffusion, weigend2023probabilistic}. Nevertheless, most of them focus on whole-body motion capture and miss fine-grained finger motions. EMF-based mocap gloves are designed for capturing finger motion, which is widely used for dexterous teleoperation~\cite{fritsche2015first, fang2015robotic, zhou2022teleman}. However, the glove does not track the 6-DoF palm poses grounded in the environment and misses visual observations for training robot policies. \sysName is a mocap glove system that is designed to collect data for training visuomotor manipulation policies. Through novel engineering designs, our system stays robust to occlusions, captures fine-grained finger motion, tracks palm poses using SLAM, and records RGB-D images to reconstruct the scene with a wearable camera vest.

\subsection{Robot learning with human demonstration}
Imitation Learning (IL) has enabled robots to successfully perform various manipulation tasks~\cite{calinon2010learning, 1014739, schaal1999imitation, kober2010imitation, englert2018learning, finn2017one, billard2008robot, argall2009survey}. Traditional IL algorithms such as DMP and PrMP~\cite{schaal2006dynamic, kober2009learning, NIPS2013_e53a0a29, paraschos2018using} enjoy high learning sample efficiency but are limited in their ability to handle high-dimensional observations. In contrast, recent IL methods built upon deep neural networks can learn policies with raw image observation inputs~\cite{mandlekar2021what, florence2019self}, even for high-degree robot systems with bimanual arms~\cite{zhao2023learning, grannen2023stabilize}. Despite their effectiveness, one key challenge for imitation learning is how to scale up the training data. Prior works focus on teleoperation data ~\cite{zhang2018deep, florence2019self, mandlekar2019scaling, ke2020telemanipulation, wang2021generalization, zhu2022viola, brohan2022rt, wu2023gello, gao2024efficient, lin2024learning} which is expensive to collect due to the requirement of the robot hardware. More recently, learning from human motion data has started to receive more attention because it allows collecting data without robot hardware~\cite{duan2023ar2}. By leveraging human videos~\cite{xu2023xskill, bharadhwaj2023towards}, hand trajectories~\cite{yang2022learning, wang2023mimicplay, gu2023rt, xiong2023robotube}, promising results have been shown to train policies with less manual human effort. However, these human motions are in 2D image space ~\cite{damen2018scaling, grauman2022ego4d, yang2022learning}, which fails to directly train 6-DoF manipulation policies in 3D environments and usually requires additional teleoperation data to bridge the gap ~\cite{xu2023xskill, bharadhwaj2023towards, wang2023mimicplay}. Recently, human-in-the-loop correction algorithms have also shown promising results in robot learning \cite{liu2022robot,peng2023learning,spencer2020learning}. Our \sysName provides tracking of 6-DoF hand poses together with finger motions grounded in 3D point cloud observations, which is portable for data collection without a robot. Based on the data collected with \sysName, we introduce \algoName which is a point cloud-based imitation learning algorithm for learning fine-grained dexterous manipulation policies, with an optional human-in-the-loop correction step for more challenging tasks.

\subsection{Portable data collection systems for manipulation}
Recently advancements in low-cost hand-held grippers have shown promising results in collecting robot manipulation data without robot hardware ~\cite{song2020grasping, young2021visual, doshi2023hand, sanches2023scalable, fang2023low, shafiullah2023bringing, chi2024universal}. All of these systems are designed and used for the parallel-gripper data collection process, while in this work we aim to collect multi-finger hand motion data for dexterous manipulation tasks (e.g., using scissors and unscrewing bottle caps).

\section{Hardware System: \sysName}
\label{sec:sys}

In this section, we introduce the system design including (1) a portable human hand motion capture system \sysName that is used for data collection (Sec. ~\ref{sec:dexcap}) and (2) a bimanual robot system equipped with dexterous hands for testing the policies learned from the collected data (Sec. ~\ref{sec:robot}).

\begin{figure*}[t]
\includegraphics[width=2.0\columnwidth]{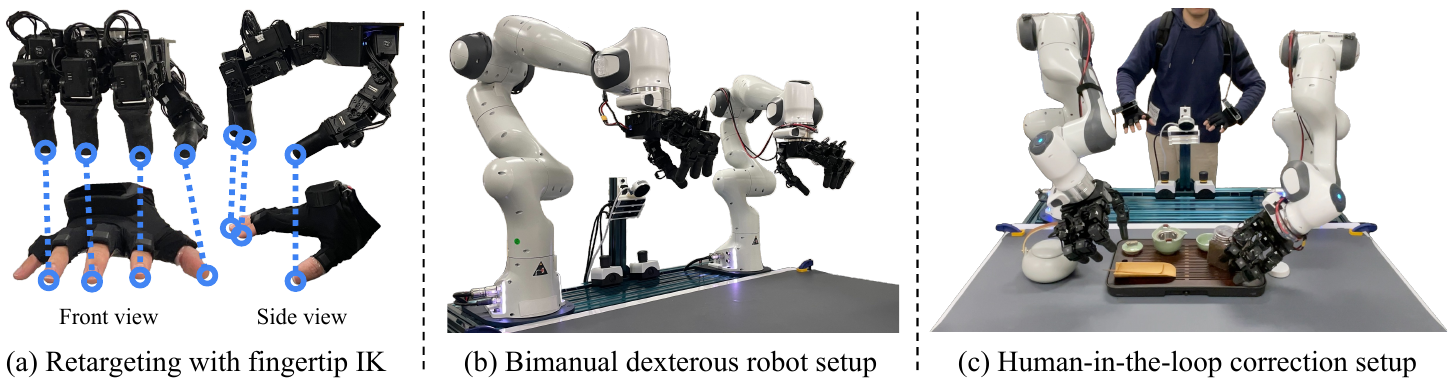}
\caption{\textbf{Details of the robot system.} Mirroring the human system, the robot system reuses the same chest cameras and mount. (a) Once the motion is captured by DexCap, it's retargeted to LEAP hand through discarding pinky finger and IK to match fingertip location. (c) An optional human-in-the-loop correction step can be performed to further refine the motions transferred. Specifically, the human will provide the delta input in real time when the robot system is carrying out the task. Note the hand T265 is only used at correction time, as the robot arm already knows the exact location of fingers.}
\vspace{-0.5cm}
\label{fig:sys_robot}
\end{figure*}

\subsection{DexCap}
\label{sec:dexcap}

To capture the fine-grained hand motion data suitable to train dexterous robot policies, \sysName is designed with four key objectives in mind: (1) detailed finger motion tracking, (2) accurate 6-DoF wrist pose estimation, (3) aligned 3D observations recording in a unified coordinate frame with hands, and (4) outstanding portability for data collection in various real-world environments. We achieved these objectives with zero compromise on \emph{scalability}---\sysName must be simple to calibrate, inexpensive to build, and robust for data collection of daily activities in the wild.

\textbf{Tracking finger motions.} Our system uses electromagnetic field (EMF) gloves, offering a significant advantage over vision-based finger tracking systems, particularly in the robustness to visual occlusions that frequently occur in hand-object interactions. In our system, finger motions are tracked using Rokoko motion capture gloves as illustrated in Figure~\ref{fig:sys}. Each glove's fingertip is embedded with a tiny magnetic sensor, while a signal receiver hub is placed on the glove's dorsal side. The 3D location of each fingertip is measured as the relative 3D translation from the hub to the sensors. In appendix we included a qualitative comparison between our EMF glove system and state-of-the-art vision-based hand-tracking methods across different manipulation scenarios.

\textbf{Tracking 6-DoF wrist pose.} Beyond finger motion, knowing the precise positioning of a robot's end-effector in a 3D space is crucial for robot manipulation. This necessitates \sysName to estimate and record the 6-DoF pose trajectories of human hands during data collection. While camera-based and IMU-based methods are commonly used, each has its limitations. Camera-based systems, often non-portable and limited in their ability to estimate wrist orientation, are less suited for data collection in manipulation tasks. IMU-based systems, although wearable, tend to suffer from position drifting when used for long recording sessions. To address these challenges, we develop a 6-DoF wrist tracking system based on the SLAM algorithm, as shown in Figure~\ref{fig:sys}(c). This system uses an Intel Realsense T265 camera, mounted on each glove's dorsal side. It combines images from two fisheye cameras and IMU sensor signals to construct an environment map using the SLAM algorithm, enabling consistent tracking of the wrist's 6-DoF pose. This design has three key advantages: it is portable, allowing for wrist pose tracking without the need for hands to be visible in third-person camera frames; SLAM can autonomously correct position drift with the built map for long-time use; and the IMU sensor provides crucial wrist orientation information to train the robot policy in the subsequent pipeline.

\begin{figure*}[t]
\includegraphics[width=2.0\columnwidth]{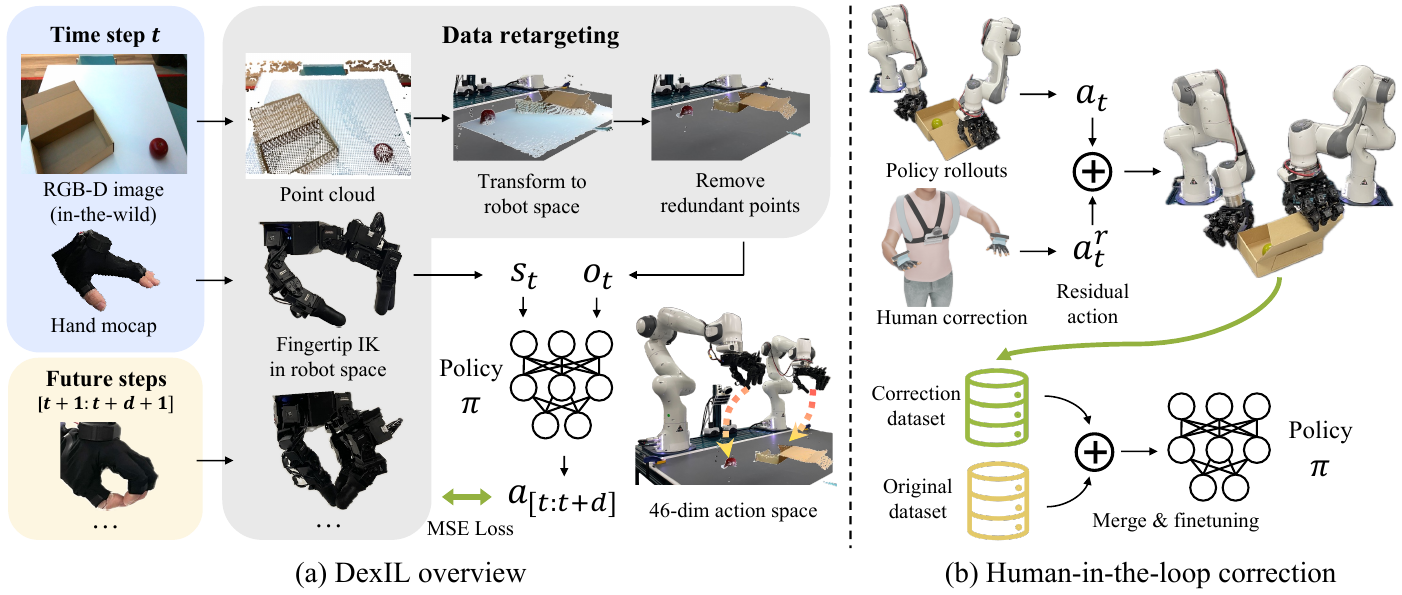}
\caption{\textbf{Algorithm overview.} (a) \algoName first retargets the \sysName data to the robot embodiment by first constructing 3D point clouds from RGB-D observations and transforming it into robot operation space. Meanwhile, the hand motion capture data is retargeted to the dexterous hand and robot arm with fingertip IK. Based on the data, a robot policy is learned to output a sequence of future goal positions as the robot actions. (b). \sysName also offers an optional human-in-the-loop correction mechanism, where humans apply delta residual action to the policy-generated actions to correct robot behavior. The corrections are stored in a new dataset and uniformly sampled with the original dataset for fine-tuning the robot policy.}
\vspace{-0.5cm}
\label{fig:algo}
\end{figure*}

\textbf{Recording 3D observations and calibration.} Capturing the data necessary for training robot policies requires not only the tracking of hand movement but also recording observations of the 3D environment as the policy input. As depicted in Figure~\ref{fig:sys}(a), we design a wearable camera vest for this purpose. It incorporates an Intel Realsense L515 RGB-D LiDAR camera, mounted on the top of the chest, to capture the observations during human data collection. The next critical question then becomes how to effectively integrate the tracked hand motion data with the 3D observations. To simplify the calibration process, we designed a 3D-printed camera rack underneath the chest camera mount as illustrated in Figure~\ref{fig:sys}(c). At the beginning of the data collection, all tracking cameras are placed in the rack slots, which secures a constant transformation between the camera frames. Then, we take off the tracking cameras from the rack and insert them into the camera slot attached to each glove. In this way, we can easily transform the hand pose tracking results into the observation frame of the chest camera with the constant initial transformation. The full calibration process is demonstrated in Appendix Figure ~\ref{fig:data_collection_prep} and supplementary videos, which takes around 10 seconds. To further ensure stable observations amidst human movement, another fisheye tracking camera (marked red in Fig.~\ref{fig:sys}(c)) is mounted under the LiDAR camera, which provides a more robust SLAM performance than the LiDAR camera with its wide field of view. We define the initial pose frame of this tracking camera as the world frame for all stream data. Figure~\ref{fig:data_vis} is the visualization of the collected data by transforming the observations into colored point clouds in the world frame alongside the captured hand motions.

\textbf{System Portability.} Central to the portability of \sysName is a compact mini-PC (Intel NUC 13 Pro), carried in a backpack, which serves as the primary computation unit for data recording. This PC is powered by a portable power bank with a 40000mAh battery, enabling approximately 40 minutes of continuous data collection (Fig.~\ref{fig:sys}(b)). The total weight of the backpack is 3.96 pounds. The supplementary video shows that donning and calibrating \sysName is fast and simple, taking less than 10 seconds. Additionally, \sysName's hardware design is modular and inexpensive to build --- no restriction to brands or models of cameras, motion capture gloves, and mini-PCs. We will open-source the code and instruction videos for builders, along with a range of hardware options. The overall cost of the \sysName is kept within a \$4k USD budget.

\subsection{Bimanual dexterous robot}
\label{sec:robot}

To validate the robot policy trained by the data from \sysName, we establish a bimanual dexterous robot setup. This setup comprises two Franka Emika robot arms, each equipped with a LEAP dexterous robotic hand (a four-fingered hand with 16 joints)~\citep{Shaw-RSS-23}, as depicted in Figure~\ref{fig:sys_robot}(b). For policy evaluation, the chest LiDAR camera used in human data collection is detached from the vest and mounted on a stand positioned between the robot arms. To simplify the process of switching the camera system between the human and robot, a quick-release buckle has been integrated into the back of the camera rack, allowing for swift camera swaps – in less than 5 seconds. In this way, the robot utilizes the same observation camera employed during human data collection. Note that, for robot setups, only the LiDAR camera is used and wrist cameras are not needed. Both the robot arms and the LEAP hands operate at a control frequency of 20Hz. We use end-effector position control for both robot arms and joint position control for both LEAP hands.

\section{Learning Algorithm: \algoName}

Our goal is to use the human hand motion capture data recorded by \sysName to train dexterous robot policies. There are several research questions along the way - (1) How can we re-target the human hand motion to the robotic hand? (2) What algorithm can learn dexterous policies, especially when the action space is high-dimensional in the bimanual setup? (3) In addition, we would like to investigate the failure cases for learning directly from human motion capture data and their potential solutions.

To tackle these challenges, we introduce \algoName, a three-step framework to train dexterous robots using human hand motion capture data. The first step is to re-target the \sysName data into the action and observation spaces of the robot embodiment (Sec.~\ref{sec:data_proc}). Second step trains a point-cloud-based diffusion policy using the re-targeted data (Sec.~\ref{sec:method}). The final step involves an optional human-in-the-loop correction mechanism, designed to address unexpected behaviors that emerge during the policy execution (Sec.~\ref{sec:human}). 

\subsection{Data re-targeting}
\label{sec:data_proc}

\textbf{Action re-targeting.} As illustrated in Figure~\ref{fig:sys_robot}(a), a notable challenge emerges due to the size disparity between the human hand and the LEAP hand, with the latter about 50\% larger~\cite{Shaw-RSS-23}. This size difference makes it hard to directly transfer the finger motions to the robotic hardware. The first step is to retarget the human hand motion capture data into the robot embodiment, which requires mapping the finger position and 6-DoF palm pose with inverse kinematics (IK). 

One critical finding in prior research is that fingertips are the most frequently contacted areas on a hand when interacting with objects (as evidenced in studies like HO-3D~\cite{hampali2020honnotate}, GRAB~\cite{taheri2020grab}, ARCTIC~\cite{fan2023arctic}). Motivated by this, we re-target finger motion by matching fingertip positions using inverse kinematics (IK). Specifically, we deploy an IK algorithm that generates smooth and accurate fingertip motion in real time~\cite{Rakita-RSS-18, rakita2021collisionik, wang2023rangedik} to determine the 16-dimensional joint positions for the robotic hand. This ensures the alignment between robot fingertips and the human fingertips in the \sysName data. Considering the design of the LEAP hand, which features four fingers, we adapt our process by excluding little finger information during IK computations. Additionally, the 6-DoF wrist pose captured in the mocap data serves as an initial reference for wrist pose in the IK algorithm. Figure~\ref{fig:data_vis} demonstrates the final result of re-targeting. The 6-DoF pose of the wrist $\bm{p}_t = [\bm{R}_t|\bm{T}_t]$ and the finger joint positions $\bm{J}_t$ of the LEAP hands are then used as the robot's proprioception state $\bm{s}_t = (\bm{p}_t, \bm{J}_t)$. We use position control in our setup and the robot's action labels are defined as next future states $\bm{a}_t = \bm{s}_{t+1}$.

\textbf{Observation post-processing.} Observation and state representation choice are critical for training robot policies. We convert the RGB-D images captured by the LiDAR camera in the \sysName data into point clouds using the camera parameters. This additional conversion offers two significant benefits compared to RGB-D input. First, because \sysName allows the human torso to move naturally during data acquisition, directly using RGB-D input would need to account for the moving camera frame. By transforming point cloud observations into a consistent world frame---defined as the coordinate frame of the main SLAM camera at the start of the mocap (the main camera is marked in red in Fig.~\ref{fig:sys}(c))---we isolate and remove torso movements, resulting in a stable robot observation. Second, point clouds provide flexibility in editing and alignment with the robot's operational space. Given that some motions captured in the wild may extend beyond the robot's reachability, adjusting the placement of point cloud observations and motion trajectories ensures their feasibility within the robot's operational range. Based on these findings, all RGB-D frames from the mocap data are processed into point clouds aligned with the robot's space, and the task-irrelevant elements, such as the table surface points, are excluded. This refined point cloud data thus becomes the observation inputs $\bm{o}_t$ fed into the robot policy $\pi$.

\subsection{Point cloud-based diffusion policy}
\label{sec:method}

With the transformed robot's state $\bm{s}_t$, action $\bm{a}_t$ and corresponding 3D point cloud observation $\bm{o}_t$, we formalize the robot policy learning process as a trajectory generation task. More specifically, a policy model $\pi$, processes the point cloud observations $\bm{o}_t$ and the robot's current proprioception state $\bm{s}_t$ into an action trajectory $(\bm{a}_t, \bm{a}_{t+1}, \dots, \bm{a}_{t+d})$ (as in Fig.~\ref{fig:algo}). Given point cloud observation with $N$ points $\bm{o}_t$ in $\mathbb{R}^{N \times 3}$, we uniformly down-sample it into $K$ points and concatenate the RGB pixel color corresponding to each point into the final policy input in $\mathbb{R}^{K \times 6}$. To bridge the visual gap between human hands and the robot's hand, we use forward kinematics to transform the links of the robot model with the proprioception state $\bm{s}_t$ and merge the point clouds of the transformed links into the observation $\bm{o}_t$. During training, we also use data augmentation over the inputs by applying random 2D translations to the point clouds and motion trajectories within the robot's operational space.

One challenge of learning dexterous robot policies, especially for bimanual dexterous robots, is handling the large dimensional action outputs. In our setup, the action output includes two 7-DoF robot arms and two 16-DoF dexterous hands for $d$ steps, which forms a high-dimensional regression problem. Similar challenges have also been studied in image generation tasks, which aim to regress all pixel values in a high-resolution frame. Recently, diffusion model~\cite{sohl2015deep, ho2020denoising}, with its step-by-step diffusion process, has shown success in modeling complex data distributions with high-dimensional data. For robotics, diffusion policy~\cite{chi2023diffusion} follows the same idea and formalizes the control problem into an action generation task. Thus we use a diffusion policy as the action decoder, where we empirically find it outperforms traditional MLP-based architecture for learning dexterous robot policies. 

\subsection{Human-in-the-loop correction}
\label{sec:human}

With the design presented above, \algoName can learn challenging dexterous manipulation skills (e.g., pick-and-place and bimanual coordination) directly from \sysName data without the need for on-robot data. However, our simple retargeting method does not address all aspects of the human-robot embodiment gap. For example, when using a pair of scissors, a stable hold of scissors requires inserting the fingers deep into the handle. Due to the differences in finger length proportion, directly matching the fingertips and the joint motion does not guarantee the same force exerted on the scissors. 

To address this issue, we offer a human-in-the-loop motion correction mechanism, which consists of two modes - residual correction and teleoperation. During policy execution, we allow humans to provide corrective actions to robots in real-time by wearing \sysName. In residual mode, \sysName measures the delta position changes of human hands $(\Delta \bm{p}_t^H, \Delta \bm{J}_t^H)$ relative to hands' initial states $(\bm{p}_0^H, \bm{J}_0^H)$ at the beginning of the policy roll-out. The delta position is applied as a residual action $\bm{a}^r_t = (\Delta \bm{p}_t^H, \Delta \bm{J}_t^H)$ to the robot policy action $\bm{a}_t = (\bm{p}_{t+1}, \bm{J}_{t+1})$, scaled by $\alpha$ and $\beta$. The corrected robot action can then be formalized as $\bm{a}'_t = (\bm{p}_{t+1} \bigoplus \alpha \cdot \Delta \bm{p}_t^H, \bm{J}_{t+1} + \beta \cdot \Delta \bm{J}_t^H)$. We empirically find that setting $\beta$ with a small scale ($<0.1$) offers the best user experience, which avoids fingers moving too fast. 

In the case when a large position change is desired, a press on the foot pedal will switch the system to teleoperation mode. \sysName now ignores the policy rollout and applies human wrist delta directly to the robot wrist pose. The robot fingertips are now directly following human fingertips. In other words, the robot fingertip will track the human fingertip in their respective wrist frame through IK. Users can also switch back to the residual mode after correcting the robot's mistake by pressing the foot pedal again.

Since the robot has already learned an initial policy, typically the correction happens in a small portion of the rollout, greatly reducing the human effort. The corrected actions and observations are stored in a new dataset $\mathcal{D}'$. Training data is sampled with equal probability from $\mathcal{D}'$ and the original dataset $\mathcal{D}$ to fine-tune the policy model, similar to IWR~\cite{mandlekar2020human}.

\begin{figure*}[t]
\includegraphics[width=2.0\columnwidth]{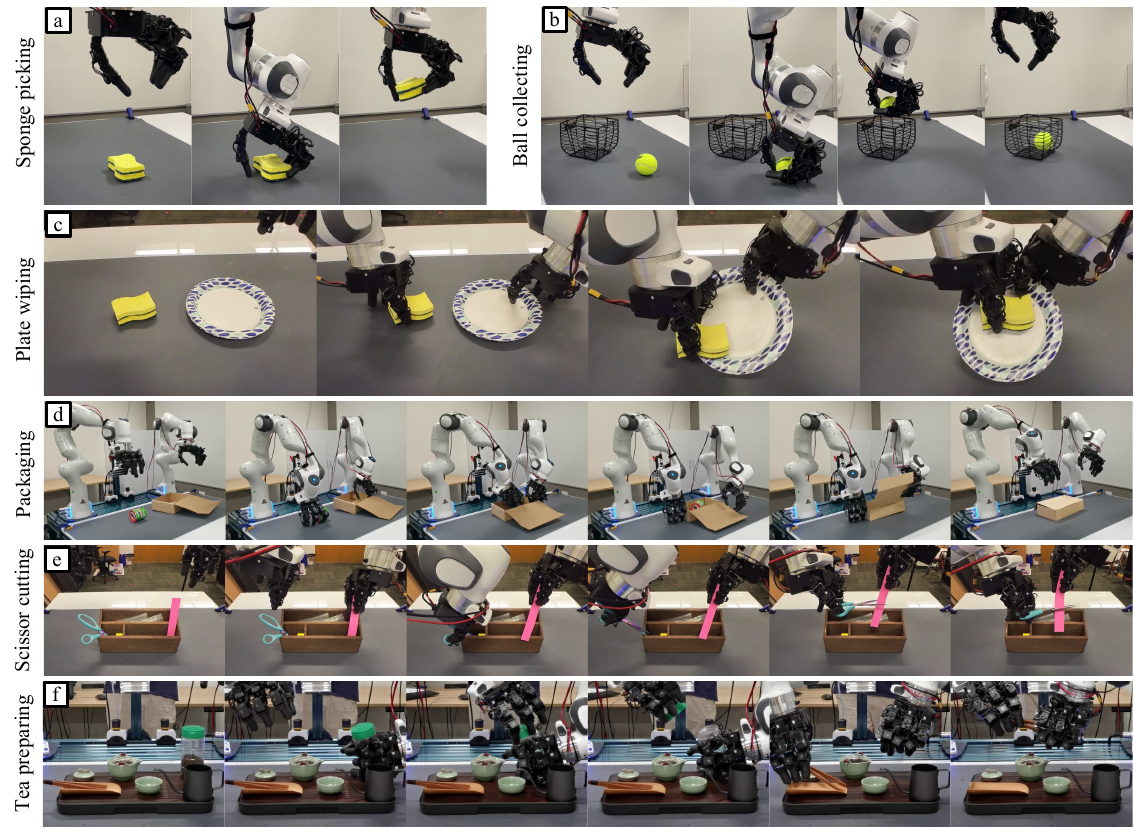}
\caption{\textbf{Experiment Tasks.} (a) \textit{Sponge Picking}: Pick and lift the sponge. (b) \textit{Ball Collecting}: Pick up a ball and drop it into a basket. (c) \textit{Plate Wiping}: Use both hands to pick up a plate and sponge, then wipe the plate vertically twice. (d) \textit{Packaging}: Place items into a box with one hand while using the other to either push or stabilize them, before securely closing the box lid. (e) \textit{Scissor Cutting}: Secure paper with one hand and use scissors in the other to cut through the paper. (f) \textit{Tea Preparing}: Grasp the tea bottle with one hand, use the other hand to uncap, then pick up tweezers to extract tea and pour it into the pot.}
\vspace{-0.3cm}
\label{fig:task_vis}
\end{figure*}

\begin{figure*}[t]
\includegraphics[width=2.0\columnwidth]{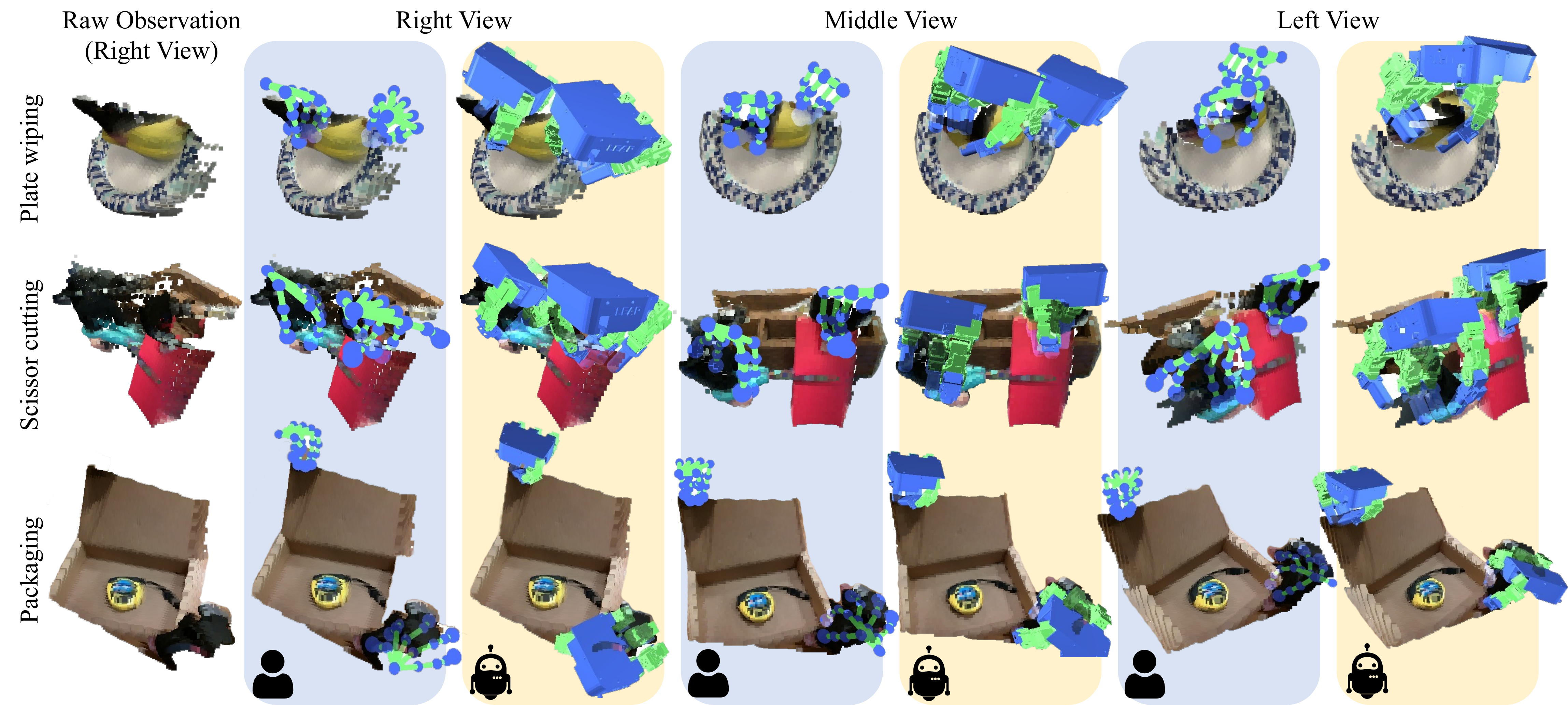}
\caption{\textbf{Data Retargeting for Tasks.} \algoName effectively retargets human mocap data for activities like plate wiping, scissor cutting, and packaging. The initial column displays the raw point cloud scene. Columns 2-7 offer three views—right, middle, left—with blue background columns depicting human data and yellow for robot hand retargeting. This side-by-side arrangement highlights the precision of our fingertip IK in translating human to robot hand motions.}
\label{fig:data_vis}
\vspace{-0.3cm}
\end{figure*}

\begin{figure}[t]
\includegraphics[width=1.0\columnwidth]{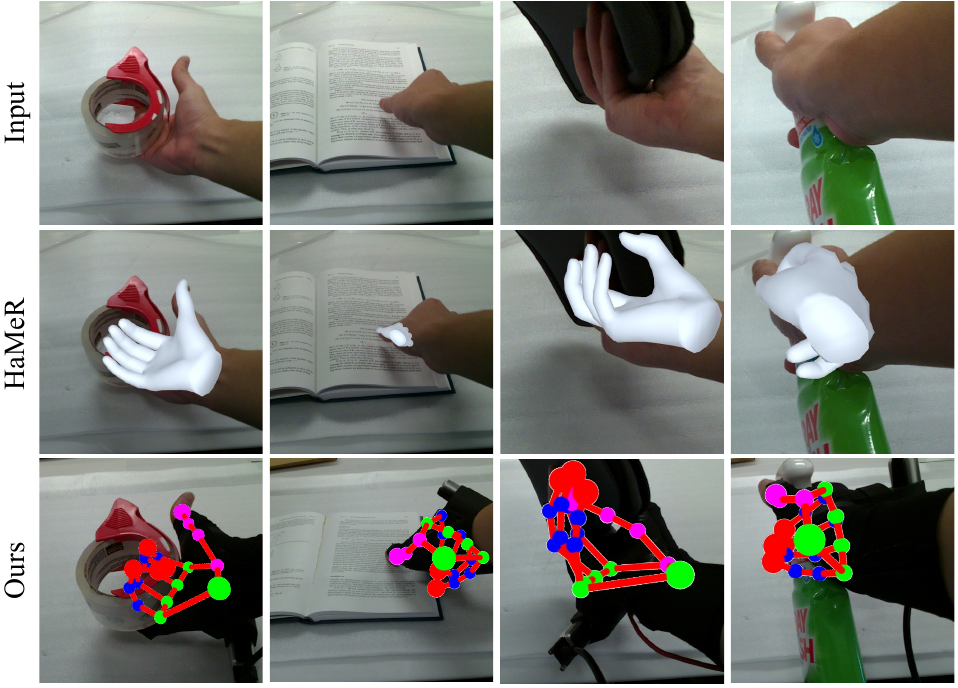}
\caption{\textbf{Compare with vision-based method.} We demonstrate that motion capture gloves provide more stable hand pose estimation results compared to vision-based methods and are not affected by visual occlusion.}
\label{fig:exp_compare}
\vspace{-0.3cm}
\end{figure}

\section{Experiments}
\label{sec:exp}

We aim to answer the following research questions: 

\begin{enumerate}
    \item[\q{1}:]{What is the quality of \sysName data?}
    \item[\q{2}:]{Can \algoName directly learn dexterous robot policies from \sysName data without any on-robot data?}
    \item[\q{3}:]{What model architecture choices are critical to improving the performance?}
    \item[\q{4}:]{Can \algoName learn from in-the-wild \sysName data?}
    \item[\q{5}:]{How does human-in-the-loop correction help when \sysName data is insufficient?}
    \item[\q{6}:]{Can the whole framework handle extremely challenging bimanual dexterous manipulation tasks (e.g., using scissors and preparing tea)?}
\end{enumerate}

\subsection{Experiment setups}

\paragraph{Tasks} we evaluate \algoName using six tasks of varying difficulty to assess its performance with \sysName data. These tasks range from basic, such as \textit{Sponge picking}, \textit{Ball collecting}, and \textit{Plate wiping}, which test single-handed and dual-handed coordination, to more complex ones like \textit{Packaging}, which looks at bimanual tasks and generalization using both familiar and new objects. \textit{Scissor cutting} focuses on the effectiveness of the human-in-the-loop correction mechanism in precise tool use, whereas \textit{Tea preparing} challenges the system with a long-horizon task requiring intricate actions. To further analyze performance, we introduce the \textbf{Subtask} metric for multi-step tasks, indicating the completion of task subgoals, such as placing an object inside a box in \textit{Packaging}, or picking up scissors in \textit{Scissor Cutting}.

\paragraph{Data} We utilize two data types: (1) \textit{\sysName data} capturing human hand motion (In-the-wild data refers to a mixture of data collected in more than 10 scenes) and (2) \textit{human-in-the-loop correction data} for adjusting robot actions or enabling teleoperation to correct errors, collected using a foot pedal. Data were initially recorded at 60Hz and then downsampled to 20Hz to match the robot's control speed, except for correction data, which was collected directly at 20Hz. For data collection, we gathered 30 minutes of \textit{\sysName data} across the first three tasks, resulting in 251, 179, and 102 demos respectively. An hour of \textit{in-the-wild \sysName data} provided 104 demos for \textit{Packaging}. \textit{Scissor Cutting} and \textit{Tea Preparing} tasks each received an hour of \sysName data, yielding 96 and 55 demos respectively.

\paragraph{Baselines} We evaluate multiple baselines to determine the model architecture with the best performance, focusing on three key aspects using \sysName data: identifying the best imitation learning framework for bimanual dexterous manipulation between BC-RNN~\cite{mandlekar2021matters} and diffusion policy (DP)\cite{chi2023diffusion}, assessing the most effective observation type to bridge the visual gap between human and robot hands (comparing image inputs~\cite{mandlekar2021what, chi2023diffusion} and a point cloud method~\cite{qin2023dexpoint}), and determining the most suitable encoder for point cloud inputs by comparing PointNet\cite{qi2016pointnet} and Perceiver~\cite{jaegle2021perceiver,lee2018set} encoders. Implementation details are included in the appendix.

\paragraph{Metric} Each model variant is tested for 20 trials in each task with randomized initial placements. The task success rate is reported in Table~\ref{tab:task123}~\ref{tab:package_task}~\ref{tab:scissor_task}. For the multi-object \textit{Packaging} task, each object is tested with 5 trials - 6 trained objects (30 total trials) and 9 unseen objects (45 total trials).

\subsection{Results}

\begin{figure*}[t]
\includegraphics[width=2.0\columnwidth]{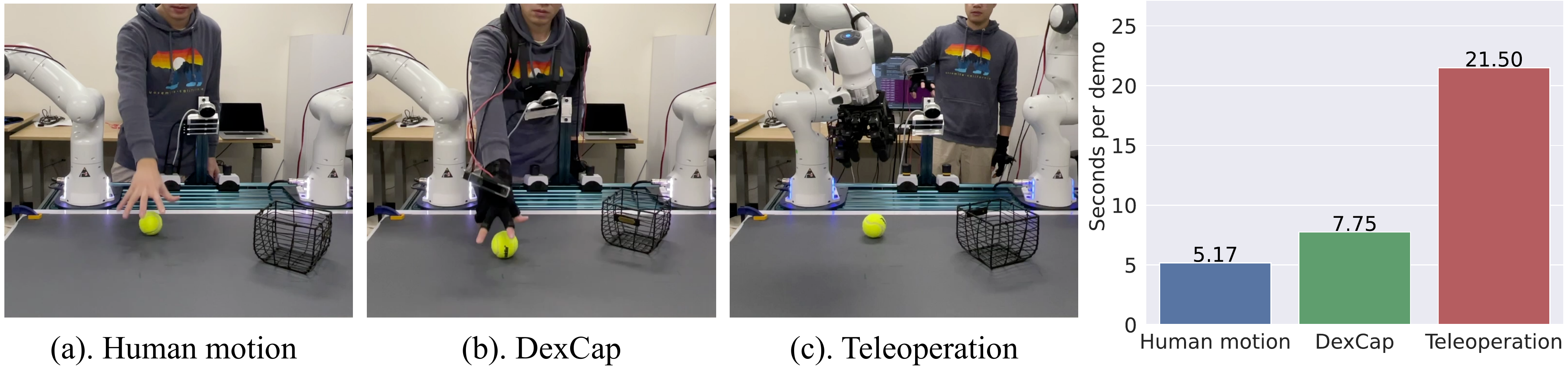}
\caption{\textbf{Data collection throughput comparison.} \sysName's data collection speed in the \textit{Ball collecting} task is close to natural human motion and is three times faster than traditional teleoperation.}
\label{fig:exp_speed}
% \vspace{-0.2cm}
\end{figure*}

\textbf{\sysName delivers high-quality 3D mocap data (\q{1}).} Figure~\ref{fig:data_vis} showcases \sysName's ability to capture detailed hand motion in 3D, aligning human actions with object point clouds across all views, such as in \textit{Plate wiping} and \textit{Scissor cutting} tasks (blue columns). The retargeted robot hand motions, depicted in the yellow columns, demonstrate precise alignment in the same 3D space. In Figure~\ref{fig:exp_compare}, we compare \sysName with the state-of-the-art vision-based hand pose estimation method HaMeR~\cite{pavlakos2023reconstructing}, observing their performance from similar viewpoints. We find that the vision-based approach is vulnerable to self-occlusion, particularly when the fingers are obscured. As depicted in Figure~\ref{fig:exp_compare}, HaMeR struggles in instances of significant occlusion, either failing to detect the hand (as seen in the second column) or inaccurately estimating fingertip positions (noted in the first, third, and fourth columns). In contrast, \sysName demonstrates good robustness under these conditions. Beyond the challenge of occlusion, most vision-based methods rely on 2D hand estimation, predicated on learning from 2D image projection losses. Consequently, these methods are inherently limited in their ability to discern the precise 3D hand positioning, as they are trained based on presumed, fixed camera intrinsic parameters, which do not necessarily match the actual camera used for experiments. In Figure~\ref{fig:exp_speed}, we showcase the data collection throughput of \sysName, which is three times faster than traditional teleoperation.

\begin{figure*}[t]
\includegraphics[width=2.0\columnwidth]{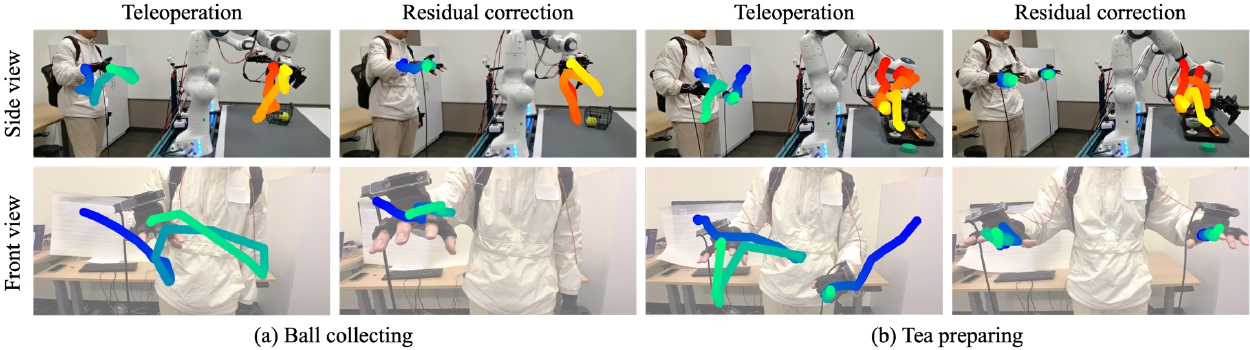}
\caption{\textbf{Visualization of human-in-the-loop corrections.} \sysName supports teleoperation and residual correction for human-in-the-loop adjustments. Teleoperation directly translates human hand movements to the robot end-effector actions, indicated by color-fading trajectories from blue to green (human) and red to yellow (robot) over 20 timesteps. Residual correction adjusts the robot's end-effector based on changes from the human hand's initial pose, enabling minimal movement but requiring more precise control. Users can switch between correction modes with a foot pedal.}
\vspace{-0.3cm}
\label{fig:exp_comp_teleop}
\end{figure*}

\textbf{\sysName data can directly train dexterous robot policies (\q{2}).} Table~\ref{tab:task123} is the experiment result of training robot policies only using \sysName data. Within 30-minute hand motion capture demonstrations collected by \sysName, the learned policies achieve up to 72\% average task success rate in single-hand pick-and-place (\textit{Sponge picking}, \textit{Ball collecting}) and bimanual coordination (\textit{Plate wiping}) tasks. This result highlights the effectiveness of \sysName data on training dexterous robot policies without on-robot data, which introduces a new way for training robot dexterous manipulation.

\textbf{Generative-based algorithm with point cloud inputs shows advantages (\q{3}).} In Table~\ref{tab:task123}, we compare the performance of multiple model architectures. We first observe that, due to the visual appearance gap between human and robot hands, the policies with full image inputs fail completely (BC-RNN-img, DP-img). We then try masking out human and robot hands with white circles in training and evaluation. This setting brings improvements, where DP-img-mask achieves more than 30\% success rate in all tasks. Meanwhile, diffusion policy works better than MLP-based BC-RNN policies (25\% higher in averaged task success rate). This result verifies our hypothesis that generative-based policy is more suitable for learning dexterous policies. Although getting promising results, masking out the end-effector loses details for in-hand manipulation. This hypothesis is verified by the low success rate in the \textit{Plate wiping} task, which requires the robot to use fine-grained finger motion to grab the plate from the edge. Our point cloud-based learning algorithms (DP-point-raw, DP-point, DP-prec), on the other hand, do not require masking over observations and achieve more than 60\% task success rate. This result highlights the advantage of using point cloud inputs, which allow us to add robot hand points to the observation without losing the details in the original inputs. We also observe that, even without adding robot hand points, DP-point-raw achieves close performance to DP-point. This might because the downsampling process of the point cloud inputs lowers the appearance gap between human gloves and robot hands. Furthermore, compared to the PointNet, the model with Perceiver encoder has higher performance, especially in bimanual tasks with multiple task objects (20\% improvement on task success rate in \textit{Plate wiping}). Based on these findings, we use DP-perc as the default model architecture for \algoName. 

\begin{table}[t]
\centering
\resizebox{\linewidth}{!}{
\footnotesize
\setlength{\tabcolsep}{0.8mm}{
\begin{tabular}{l|ccc|c}
\toprule
              & \multicolumn{4}{c}{\sysName Data Only} \\
                \cline{2-5}
              \\[-1.8ex]
              & Sponge picking & Ball collecting & Plate wiping & Overall \\
\midrule
BC-RNN-img        & 0.00 & 0.00 & 0.00 & 0.00 \\
BC-RNN-img-mask~\cite{mandlekar2021what}   & 0.25 & 0.10 & 0.10 & 0.15 \\
BC-RNN-point~\cite{qin2023dexpoint}   & 0.45 & 0.30 & 0.25 & 0.33 \\
BC-RNN-prec~\cite{jaegle2021perceiver}      & 0.50 & 0.30 & 0.35 & 0.38 \\
DP-img            & 0.00 & 0.00 & 0.00 & 0.00 \\
DP-img-mask~\cite{chi2023diffusion}       & 0.55 & 0.40 & 0.30 & 0.42 \\
DP-point-raw       & 0.70 & \textbf{0.70} & 0.40 & 0.60 \\
DP-point       & 0.75 & 0.65 & 0.50 & 0.63 \\
Ours (DP-perc)     & \textbf{0.85} & 0.60 & \textbf{0.70} & \textbf{0.72} \\
\bottomrule    
\end{tabular}
}}
\caption{Quantitative results for learning with \sysName data.}
\vspace{-0.3cm}
\label{tab:task123}
\end{table}

\begin{table}[t]
\centering
\resizebox{\linewidth}{!}{
\footnotesize
\setlength{\tabcolsep}{2pt}{
\begin{tabular}{l|ccc|ccc}
\toprule
\multicolumn{1}{c|}{\multirow{2}{*}{Packaging}} & \multicolumn{3}{c|}{In-the-wild \sysName} & \multicolumn{3}{c}{\makebox[0pt][c]{30 human corrections}} \\
              & \multicolumn{1}{c}{Subtask} & \multicolumn{1}{c}{All} & \multicolumn{1}{c|}{Unseen} & \multicolumn{1}{c}{Subtask} & \multicolumn{1}{c}{All} & \multicolumn{1}{c}{Unseen} \\
\midrule
BC-RNN-img-mask~\cite{mandlekar2021what} & 0.00 & 0.00 & 0.00& 0.23 & 0.07 & 0.00 \\
BC-RNN-point~\cite{qin2023dexpoint}      & 0.33 & 0.23 & 0.16 & 0.40 & 0.27 & 0.22 \\
DP-img-mask~\cite{chi2023diffusion}     & 0.17 & 0.00 & 0.00 & 0.47 & 0.33 & 0.00 \\
Ours          & \textbf{0.70} & \textbf{0.47} & \textbf{0.40} & \textbf{0.83} & \textbf{0.57} & \textbf{0.42} \\
\bottomrule    
\end{tabular}
}}
\caption{Quantitative results for the \textit{Packaging} task.}
\vspace{-0.3cm}
\label{tab:package_task}
\end{table}

\begin{table}[t]
\centering
\resizebox{\linewidth}{!}{
\footnotesize
\setlength{\tabcolsep}{4.0mm}{
\begin{tabular}{l|cc|cc}
\toprule
\multicolumn{1}{c|}{\multirow{2}{*}{Scissor cutting}} & \multicolumn{2}{c|}{\sysName Data Only} & \multicolumn{2}{c}{30 human corrections} \\
              & \multicolumn{1}{c}{Subtask} & All & Subtask & All \\
\midrule
BC-RNN-point~\cite{qin2023dexpoint}      & 0.00 & 0.00 & 0.10 & 0.00 \\
Ours          & 0.00 & 0.00 & \textbf{0.45} & \textbf{0.20} \\
\bottomrule    
\end{tabular}
}}
\caption{Quantitative results for the \textit{Scissor cutting} task.}
\vspace{-0.3cm}
\label{tab:scissor_task}
\end{table}

\begin{table}[t]
\centering
\resizebox{\linewidth}{!}{
\footnotesize
\setlength{\tabcolsep}{4.0mm}{
\begin{tabular}{l|cc|cc}
\toprule
\multicolumn{1}{c|}{\multirow{2}{*}{Tea preparing}} & \multicolumn{2}{c|}{\sysName Data Only} & \multicolumn{2}{c}{30 human corrections} \\
              & \multicolumn{1}{c}{Subtask} & All & Subtask & All \\
\midrule
Ours          & 0.30 & 0.00 & 0.65 & 0.25 \\
\bottomrule    
\end{tabular}
}}
\caption{Quantitative results for the \textit{Tea preparing} task.}
\label{tab:tea_task}
\vspace{-0.3cm}
\end{table}

\begin{figure}[t!]
\includegraphics[width=1.0\columnwidth]{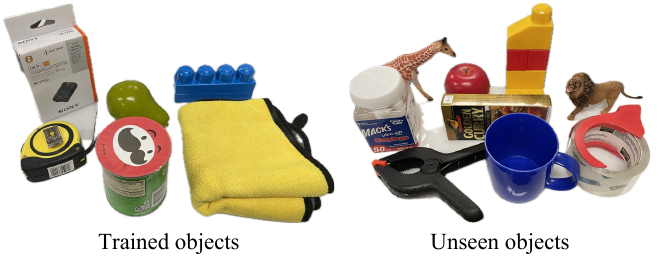}
\vspace{-0.5cm}
\caption{\textbf{Objects used in the \textit{Packaging} task} 
}
\label{fig:exp_obj_vis}
\vspace{-0.3cm}
\end{figure}

\textbf{\algoName can purely learn from in-the-wild \sysName data (\q{4}).} The first three columns of Table~\ref{tab:package_task} are the results of training policies using in-the-wild \sysName data. We first notice that image-input baselines (BC-RNN-img-mask, DP-img-mask) have close to zero performance when learning with in-the-wild data. This observation verifies our hypothesis that the viewpoint changes caused by human body movements during in-the-wild data collection bring challenges to learning image-based policies. Our \algoName transforms the point cloud inputs into a consistent world frame, resulting in stable observations and thus getting better results (70\% in Subtask and 47\% in full task setup). Please refer to our video results for more visualization of the stabilized input point clouds. By training the policy with multiple task objects using in-the-wild (Fig.~\ref{fig:exp_obj_vis}), our model can already generalize to unseen object instances, with a 40\% success rate. During evaluation, we identified two primary issues with the policy learned from in-the-wild \sysName data: firstly, the absence of force information in \sysName data causes the right hand to struggle with stabilizing the box during box closure attempts by the left hand. Secondly, the box lid occasionally moves out of the chest camera's view due to human movements, hindering the robot's ability to learn precise lid grasping. These challenges prompt us to seek improvement strategies.

\textbf{Human-in-the-loop correction greatly help when \sysName data is insufficient (\q{5}).} Figure~\ref{fig:exp_comp_teleop} illustrates two types of human-in-the-loop correction mode with \sysName. Users can switch between the two modes by stepping on the foot pedal and the whole trajectory is stored and used for fine-tuning the policy. The last three columns of Table~\ref{tab:package_task} showcase the effectiveness of using human-in-the-loop correction together with policy fine-tuning to improve the model performance. With just 30 human correction trials during policy rollout, the fine-tuned policy with image inputs (DP-img-mask) achieves a 33\% improvement in the full task success rate for trained objects. This significant boost is mainly because the human correction data is collected using a fixed camera - the same setup used for the evaluations. This result further supports our conclusion: image-based approaches are more effective in learning with fixed third-view cameras compared to the in-the-wild scenarios with moving cameras. Human corrections also result in a 10\% improvement in our approach that utilizes point cloud inputs. However, we've observed that fine-tuning with human corrections has a minor effect on the results for unseen objects, primarily due to the limited amount of correction data (30 trials in total).

\textbf{Our whole framework can handle extremely challenging tasks (\q{6}).} \algoName together with human-in-the-loop correction is able to solve extremely challenging tasks such as \textit{Scissor cutting} and \textit{Tea preparing}. In Table~\ref{tab:scissor_task}, we showcase that our system can achieve a 45\% success rate on picking up the scissor from the container and 20\% in cutting a piece of paper tape. In our supplementary video, we also showcase how the robot performs the long-horizon \textit{Tea preparing} task which includes unscrewing a bottle cap and pouring tea into the pot. Table~\ref{tab:tea_task} presents the evaluation results of our approach (DP-perc) in the \textit{Tea preparing} task. The subtask is defined as successfully unscrewing the cap of the tea bottle. We found that even with human mocap data only (\sysName Data Only), our model can achieve a 30\% success rate in uncapping. Most of the failures occur during the task of picking up the tweezers, which requires high-precision control over the fingertip. In such cases, human-in-the-loop correction significantly improves performance. With 30 human corrections, we achieve a 35\% improvement in the uncapping success rate and attain a 25\% success rate for the entire task. Please refer to our video submission for more qualitative results of this task. These tasks showcase the high potential of our framework in learning extremely challenging dexterous manipulation tasks.

\section{Conclusion and Limitations}
\label{sec:conclude}

We present \sysName, a portable hand motion capture system, and \algoName, an imitation algorithm enabling robots to learn dexterous manipulation directly from human mocap data. \sysName, designed to overcome occlusions, capture fine-grained 3D hand motion, record RGB-D observations, and allow data collection outside the lab. \algoName applies this data to teach robots complex dexterous manipulation tasks, with an optional human-in-the-loop correction mechanism to further improve performance. Demonstrating proficiency in tasks like scissor cutting and tea preparation, \sysName and \algoName significantly advance robotic dexterity. We hope \sysName can pave the path for future research on scaling up dexterous manipulation data with portable devices. All hardware designs and code will be open-source.

While \mbox{\sysName} collects high-quality mocap data in-the-wild for learning challenging dexterous manipulation tasks, it has several limitations that need future research: (1) The system's power consumption currently restricts the collection time to be at most 40 minutes. Future improvements will focus on enhancing power efficiency to extend the collection time. (2) Our learning algorithm \mbox{\algoName} utilizes fingertip inverse kinematics to retarget human hand motion to various robotic hands. However, the size difference between human and robotic hands (with some robotic fingers being thicker) can make some tasks difficult to perform, such as playing the piano. Future developments will aim to integrate advancements in robotic hand design to minimize these size differences and fully demonstrate the system's potential. (3) Current \mbox{\sysName} collects only 3D observations and motion capture data, lacking force sensing. One promising direction we plan to explore involves the use of conformal tactile textiles, as introduced in \mbox{\cite{luo2021learning}}, to gather tactile information during data collection.

\section*{Acknowledgments}
This research was supported by National Science Foundation NSF-FRR-2153854 and Stanford Institute for Human-Centered Artificial Intelligence, SUHAI. This work is partially supported by ONR MURI N00014-21-1-2801. We would like to thank Yunfan Jiang, Albert Wu, Paul de La Sayette, Ruocheng Wang, Sirui Chen, Josiah Wong, Wenlong Huang, Yanjie Ze, Christopher Agia, Jingyun Yang and the SVL PAIR group for providing help and feedback. We also thank Zhenjia Xu, Cheng Chi, Yifeng Zhu for their suggestions on the robot controller. We especially thank Kenneth Shaw, Ananye Agrawal, Deepak Pathak for open-sourcing the LEAP Hand.

\bibliographystyle{unsrtnat}
\bibliography{references}

\clearpage
\clearpage
\appendices

\section{Implementation Details}

\subsection{\sysName hardware implementations}
\begin{figure}[t]
\includegraphics[width=1.0\columnwidth]{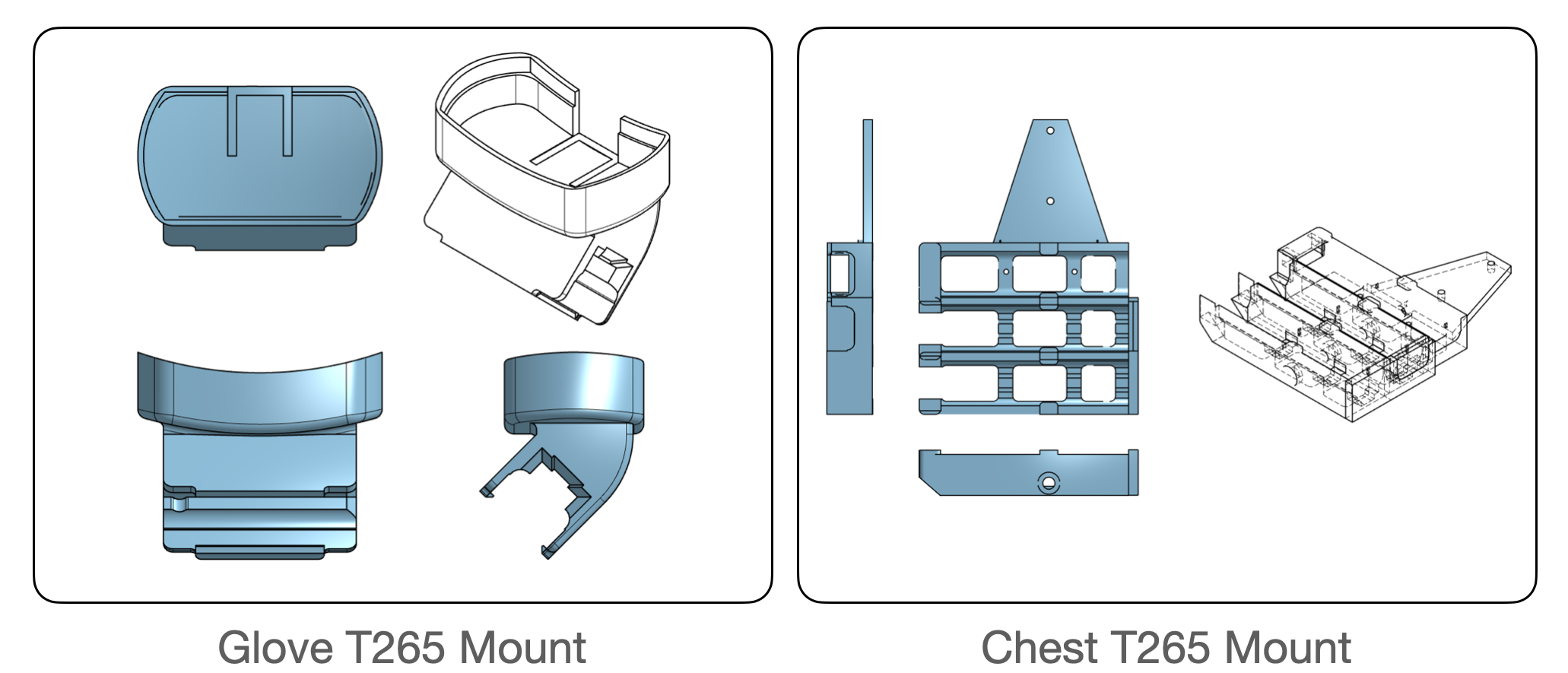}
\caption{\textbf{Detailed view of chest mount and glove mount} The glove mount follows the contour of the hump on the top of the Rokoko glove, and an opening is added to route the USB-C cable to the glove. The angle of the camera is set to 45 degrees facing upwards so that the camera view is less obstructed from the back of the hand. The slide guide has an indentation matching the position of the back plate to ensure the same insertion position across experiments. The chest mount houses 3 identical slots following the contour of the T265. An additional slot is added to fit in the slide plate of the T265.}
\label{fig:3view}
\end{figure}

Figure~\ref{fig:3view} illustrates the hardware design of \sysName. All models are 3D-printed with PLA material. The chest camera mount is equipped with four slots for cameras: at the top, an L515 RGB-D LiDAR camera, followed by three T265 fisheye SLAM tracking cameras. The LiDAR camera and the uppermost T265 camera are securely fixed to the camera rack, while the two lower T265 cameras are designed to be detachable and can be affixed to the glove's back for hand 6-DoF pose tracking. The design features of the camera mounts on both the chest and gloves include a locking mechanism to prevent the cameras from accidentally slipping out. On the glove, the camera mount is positioned over the magnetic hub on its dorsal side, ensuring a firm attachment between the hub and the mount. For powering and data storage, the user wears a backpack containing a 40000mAh portable power bank and a mini-PC with 64GB RAM and 2TB SSD. The system's total weight is 3.96 pounds, optimized for ease of mobility, supporting up to 40 minutes of continuous data collection. The power bank's rapid recharge capability, requiring only 30 minutes for a full charge, enables extensive data collection sessions over several hours.

\subsection{Data collection details}
\begin{figure*}[t]
\includegraphics[width=2.0\columnwidth]{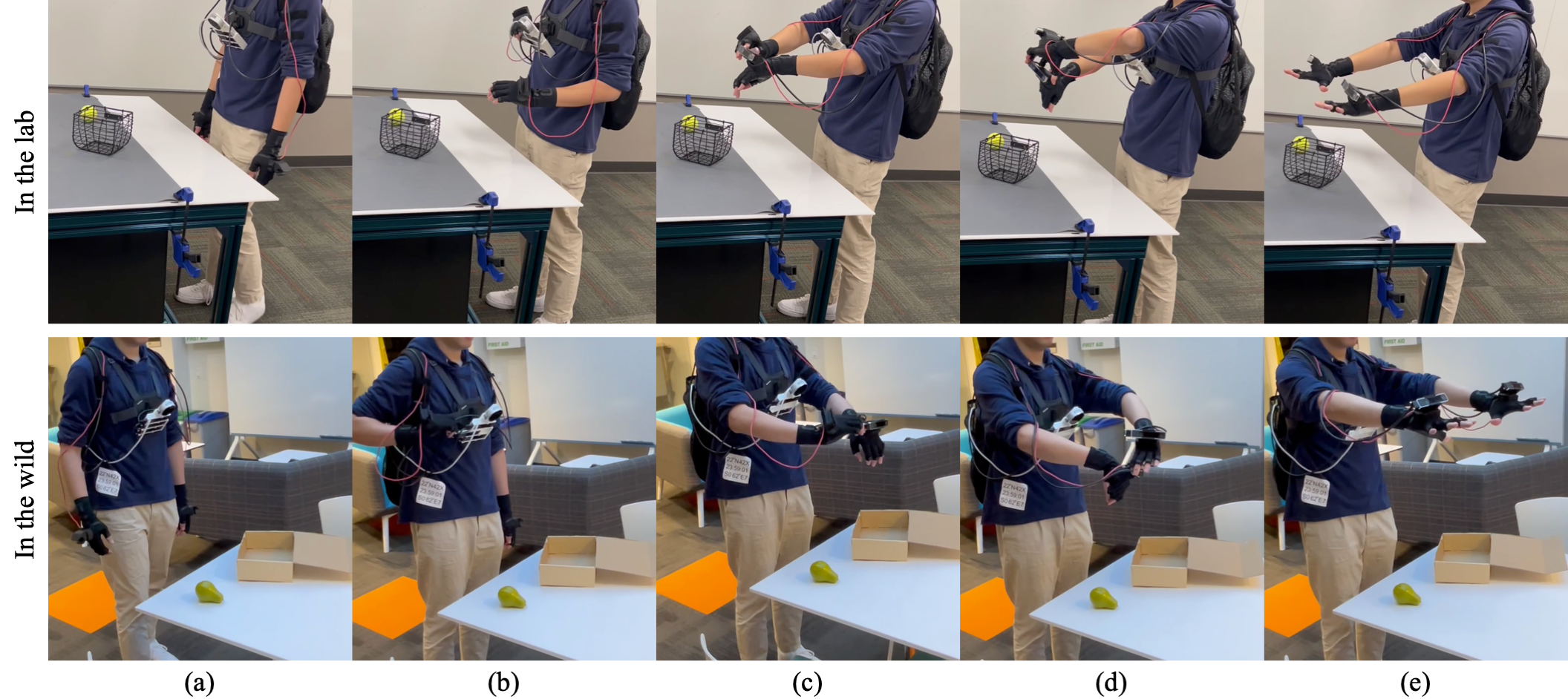}
\caption{\textbf{Prepration of data collection in the wild.} The first row illustrates data collection conducted in a laboratory setting, and the second row depicts in-the-wild data collection. (a) Initially, the human data collector moves around in the environment to track 6-DoF wrist poses with SLAM. (b)-(d) Subsequently, the data collector detaches the two cameras from the chest mount and secures them onto the glove mount. (e) With this setup, the human is prepared to begin data collection.}
\label{fig:data_collection_prep}
\end{figure*}

Figure ~\ref{fig:data_collection_prep} and the supplementary video illustrate the beginning steps of a data collection session. Initially, all cameras are mounted on the chest. Upon initiating the program, the participant moves within the environment for several seconds, allowing the SLAM algorithm to build the map of the surroundings. Subsequently, the bottom T265 cameras are relocated to the glove mounts, initiating the data collection phase. This preparatory phase is completed in approximately 15 seconds, as demonstrated in the video submission.

The data collection encompasses four data types, recorded at 60 frames per second: (1) the 6-DoF pose of the chest-mounted LiDAR camera, as tracked by the top T265 camera; (2) the 6-DoF wrist poses, as captured by the two lower T265 cameras attached to the gloves; (3) the positions of finger joints within each glove's reference frame, detected by the motion capture gloves; and (4) RGB-D image frames from the LiDAR camera. The initial pose of the top T265 camera establishes the world frame for all data, allowing for the integration of all streamed data—RGB-D point clouds, hand 6-DoF poses, and finger joint locations—into a unified world frame. This configuration permits unrestricted movement by the participant, enabling easy isolation and removal of body movements from the dataset.

Data are initially buffered in the mini-PC's RAM, supporting a 15-minute collection at peak frame rate (60 fps). Once the RAM is full, data capture slows to 20 fps due to storage shifting to the SSD. We empirically find that this reduction in frame rate may affect SLAM tracking accuracy, potentially leading to jumping tracking results. Thus, we use the first 10 minutes of each session prioritized for high-quality data capture. After collection, transferring the data from RAM to SSD is efficiently completed within 3-5 minutes using multi-threading.

In this study, we primarily investigate two types of \sysName data: (1) data captured in the robot space and (2) data collected in the wild. For the first category, we position the chest camera setup on a stand between two robot arms. The robots are then adjusted to a resting position, clearing the operational space for human interaction. This arrangement allows for the direct use of \sysName to collect data within the robot's operational area. Such data underpins basic experiments for tasks like \textit{Sponge picking}, \textit{Ball collecting}, and \textit{Plate wiping}, alongside more complex challenges, including \textit{Scissor cutting} and \textit{Tea preparing}. For the second category, individuals don \sysName to gather data outside the lab setting, focusing on the system's zero-shot learning performance with \textit{in-the-wild} \sysName data and its ability to generalize to unseen objects, particularly in the \textit{Packaging} task.

\subsection{Data retargeting details}

To adapt the collected raw \sysName data for training robot policies (commonly known as retargeting). This involves two key steps: (1) retargeting the observations and (2) retargeting the actions.

For observation retargeting, the initial step is to convert the RGB-D inputs into 3D point clouds, ensuring each pixel's color is preserved. These point clouds are then aligned with the world frame, defined by the initial pose of the main T265 camera. Subsequently, a point cloud visualization UI is launched, displaying the aligned input point clouds alongside the robot operation space's point clouds within a unified coordinate frame. Through this UI, users can adjust the point cloud's position within the robot operation space using the keyboard's directional keys. This adjustment process is required only once for all data collected in the same location and is completed in under a minute. After aligning the point clouds with the robot space, points below the robot's table surface are eliminated, refining the observation data for policy development.

\begin{figure*}[t]
\includegraphics[width=2.0\columnwidth]{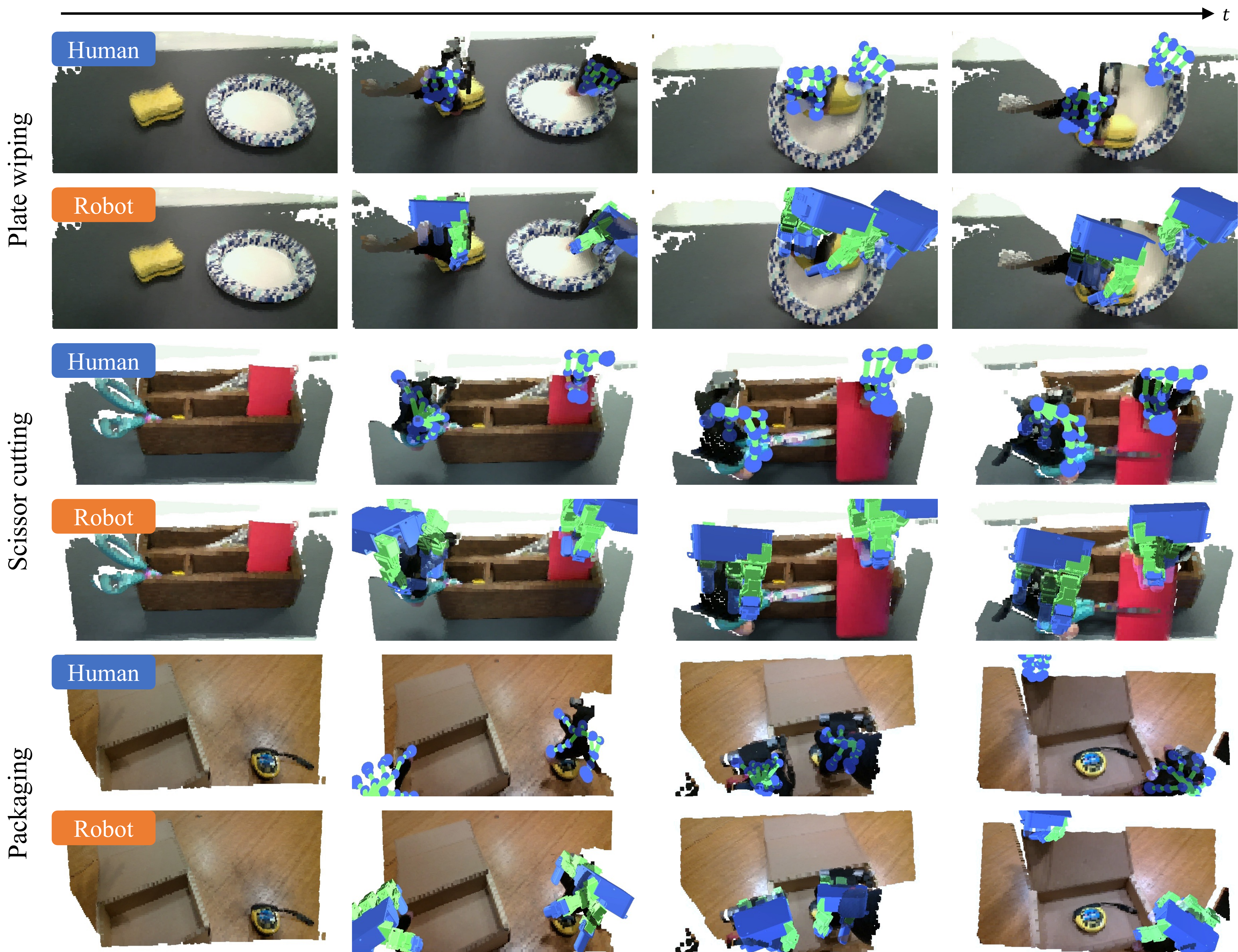}
\caption{\textbf{Visualization of collected human data and retargeted robot data.} \algoName successfully adapts human motion capture data for tasks such as plate wiping, scissor cutting, and packaging. We demonstrate the entire workflow of executing these tasks.}
\label{fig:app_data_vis}
\end{figure*}

Action retargeting begins with applying a consistent transformation between the T265 cameras on the chest mount to translate the hand joint locations into the world frame. Then, we use the previously calculated point cloud transformation matrix to transform the hand joints to the robot operation space. The results of this process are visualized in Figure~\ref{fig:app_data_vis} by depicting the transformed hand joints together with the point cloud as a skeletal model of the hand. The final phase employs inverse kinematics to map the fingertip positions between the robot hand (LEAP hand) and the human hand. We use the hand's 6-DoF pose to initialize the LEAP hand's orientation for IK calculation. Figure~\ref{fig:app_data_vis} illustrates the IK results, showing the robot hand model integrated with the observational point clouds, thereby generating the actions required for training the robot policy.

All of the point cloud observations are downsampled uniformly to 5000 points and stored together with robot proprioception states and actions into an hdf5 file. We manually annotate the start and end frames of each task demonstration from the entire recording session (10 minutes each). The motion for resetting the task environment is not included in the training dataset.

\begin{figure*}[t]
\includegraphics[width=2.0\columnwidth]{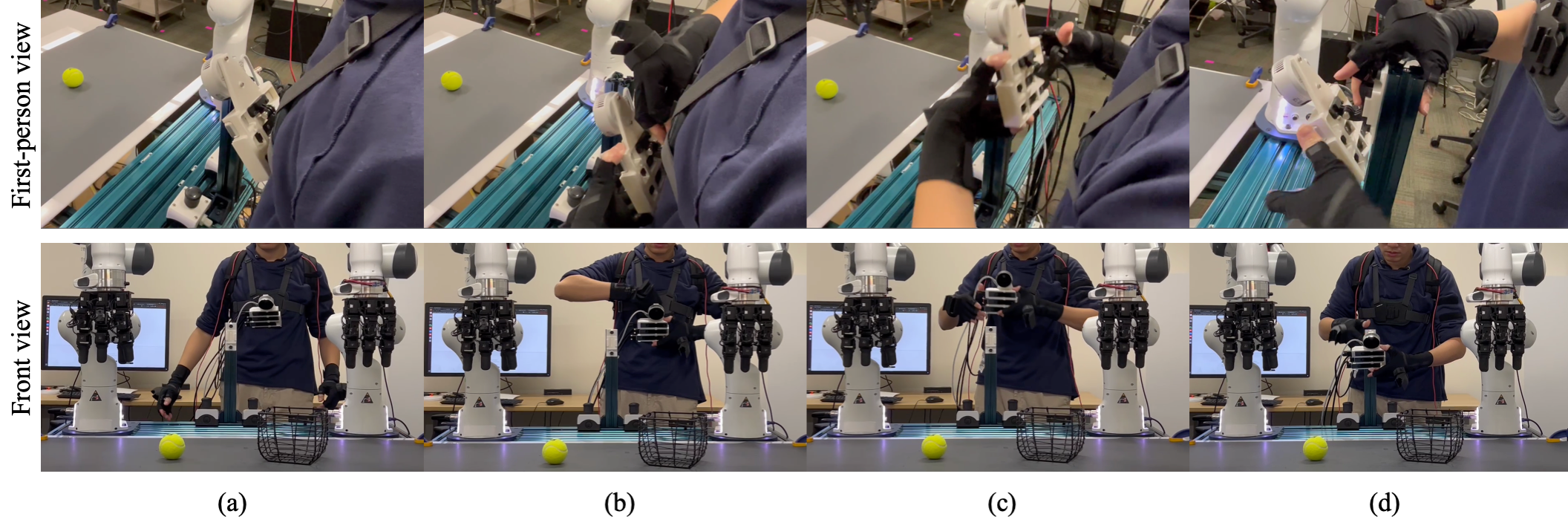}
\caption{\textbf{Switching \sysName from the human to the robot.} We illustrate, from both first-person and front views, the seamless transition of \sysName from a human data collector to a bimanual dexterous robot system. This process involves effortlessly detaching the cameras from the chest mount and inserting them into a stationary mount on the robot's table.}
\label{fig:system_switch}
\end{figure*}

\subsection{Robot controller details}

Position control is employed throughout our experiments, structured hierarchically: (1) At the high level, the learned policy generates the goal position for the next step, which encompasses the 6-DoF pose of the end-effector for both robot arms and a 16-dimensional finger joint position for both hands. (2) At the low level, an Operational Space Controller (OSC)~\cite{khatib1987unified}, continuously interpolates the arm's trajectory towards the high-level specified goal position and relays interpolated OSC actions to the robot for execution. Meanwhile, finger movements are directly managed by a joint impedance controller. Following each robot action, we calculate the distance between the robot's current proprioception and the target pose. If the distance between them is smaller than a threshold, we regard that the robot has reached the goal position and will query the policy for the next action. To prevent the robot from becoming idle, if it fails to reach the goal pose within $h$ steps, the policy is queried anew for the subsequent action. We designate $h = 10$ in our experiments. We empirically find that for tasks that consist of physical contact with objects or applying force, this situation happens more often and a smaller $h$ will have a smoother robot motion.

\subsection{Policy model and training details}
For all image-input methods, we use ResNet-18~\cite{he2016deep} as the image encoder. For models based on diffusion policy, we use Denoising Diffusion Implicit Models (DDIM)~\cite{song2020denoising} for the denoising iterations. For all baselines, the time horizon of the inputs is set to three. For pointcloud-based methods, the input point cloud is uniformly downsampled to 1000 points. We list the hyperparameters for each architecture in Table~\ref{tab:hyperparameters_bc_rnn_img}, ~\ref{tab:hyperparameters_dp_point}, ~\ref{tab:hyperparameters_ours}.

\begin{table}[t]
  \makeatletter\def\@captype{table}
    \centering
\small
\resizebox{0.5\linewidth}{!}{
\centering
    \begin{tabular}{cc}
    \toprule
    Hyperparameter & Default \\
    \midrule
      Batch Size & 16 \\
      Learning Rate (LR)   &  1e-4 \\
      Num Epoch & 3000 \\
      LR Decay & None \\
      Image Encoder & ResNet-18 \\
      Image Feature Dim & 64 \\
      RNN Type & LSTM \\
      RNN Horizon & 3 \\
      GMM & None \\
    \bottomrule
    \end{tabular}
}
\caption{Hyperparameters - BC-RNN-img}
\label{tab:hyperparameters_bc_rnn_img}
\end{table}

\begin{table}[t]
  \makeatletter\def\@captype{table}
    \centering
\small
\resizebox{0.6\linewidth}{!}{
\centering
    \begin{tabular}{cc}
    \toprule
    Hyperparameter & Default \\
    \midrule
      Batch Size & 16 \\
      Learning Rate (LR)   &  1e-4 \\
      Num Epoch & 3000 \\
      LR Decay & None \\
      Point Cloud Encoder & PointNet \\
      Point Cloud Downsample & 1000 \\
      Pooling Type & MaxPooling \\
      UNet Embed Dim & 256 \\
      UNet Down dims & [256, 512, 1024] \\
      UNet Kernel Size & 5 \\
      Diffusion Type & DDIM \\
      Diffusion Num Train & 100 \\
      Diffusion Num Infer & 10 \\
      Input Horizon & 3 \\
    \bottomrule
    \end{tabular}
}
\caption{Hyperparameters - DP-point}
\label{tab:hyperparameters_dp_point}
\end{table}

\begin{table}[t]
  \makeatletter\def\@captype{table}
    \centering
\small
\resizebox{0.6\linewidth}{!}{
\centering
    \begin{tabular}{cc}
    \toprule
    Hyperparameter & Default \\
    \midrule
      Batch Size & 16 \\
      Learning Rate (LR)   &  1e-4 \\
      Num Epoch & 3000 \\
      LR Decay & None \\
      Point Cloud Encoder & Perceiver \\
      Point Cloud Downsample & 1000 \\
      Pooling Type & MaxPooling \\
      UNet Embed Dim & 256 \\
      UNet Down dims & [256, 512, 1024] \\
      UNet Kernel Size & 5 \\
      Diffusion Type & DDIM \\
      Diffusion Num Train & 100 \\
      Diffusion Num Infer & 10 \\
      Input Horizon & 3 \\
    \bottomrule
    \end{tabular}
}
\caption{Hyperparameters - Ours (DP-prec)}
\label{tab:hyperparameters_ours}
\end{table}

\subsection{Task implementations}
In this section, we introduce the details of each task design

\begin{itemize}
    \item \textit{Sponge Picking}: A sponge is randomly placed on the table within a 40$\times$70 centimeter area. The objective is to grasp the sponge and lift it upwards by more than 30 centimeters.
    \item \textit{Ball Collecting}: A ball is randomly positioned on the right side of the table within a 40$\times$30 centimeter area, while a basket is similarly placed randomly on the left side within the same dimensions. The task is completed when the ball is grasped and then dropped into the basket.
    \item \textit{Plate Wiping}: In a setup akin to the \textit{Ball Collecting} task, a plate and a sponge are randomly placed on the right and left sides of the table, respectively, each within a 40$\times$30 centimeter area. The goal involves using both hands to pick up the plate and sponge separately, then utilizing the sponge to wipe the plate twice. This task demands coordination between the two hands, positioning the plate in the table's middle area to facilitate the wiping action.
    \item \textit{Packaging}: An empty paper box and a target object are randomly positioned on the table, with the object within a 40$\times$30 centimeter area on the right and the box within a 10$\times$10 centimeter area on the left. This task aims to assess the model's ability to generalize across various objects, including unseen ones not present in the training dataset. Success involves using one hand to pick up the object and the other to move the box to the table's center. The object is then placed into the box, followed by stabilizing the box with one hand while the other closes it by grasping and moving the lid.
    \item \textit{Scissor Cutting}: A container is fixed at the table's center, with scissors on the left and a strip of paper tape on the right. The task begins with the left hand functionally grasping the scissors—inserting the thumb into one handle and the index and middle fingers into the other. Simultaneously, the right hand grasps the paper tape. Both scissors and tape are then lifted and moved towards the center, with the left hand operating the scissors to cut the tape. A cut exceeding 3 millimeters deems the task successful.
    \item \textit{Tea Preparing}: A tea table is centrally placed with a fixed orientation, accompanied by a tea bottle, tweezers, and a teapot. The robot must first grasp the tea bottle with the left hand and unscrew the cap with the right hand, completing two rotations. The cap is then taken off and placed on the right side of the tea table. Subsequently, the right hand picks up the tweezers from the top right corner of the tea table. The robot then attempts to pour tea from the bottle into the teapot with the left hand, while the right hand uses the tweezers to aid the pouring process. Finally, the robot returns the tweezers and the tea bottle to their corresponding positions on the table. The task is deemed successful if tea makes it into the teapot and both the tea bottle and tweezers are returned to their respective places. For the task to be considered fully successful, the tea bottle must be completely released from the left hand.
\end{itemize}

\subsection{Human-in-the-loop implementations}

\sysName incorporates two human-in-the-loop correction methodologies: teleoperation and residual correction. Both methods can be utilized during policy rollouts to gather additional correction data, which is used in further refining the policy for enhanced task performance. Detailed descriptions of these algorithms and their implementation are provided in the main paper. In the human-in-the-loop process, we employ the mini-PC to live stream data from all T265 tracking cameras. This tracking information is then transmitted to a Redis server configured on the local network. Concurrently, the robot, operating the learned policy on a workstation, receives delta movements of the human hands from the Redis server. These deltas serve as residual corrections and are integrated into each robot action. The RGB-D LiDAR camera, positioned on the central bar between the robot arms, connects to the workstation to capture observation data. Instead of recording the robot's actual positional changes, we log the action commands dispatched to the robot controller. This design is crucial for tasks involving physical contact with the environment and objects.

\section{Supplementary Experiment Results}

\begin{figure}[t]
\includegraphics[width=0.95\columnwidth]{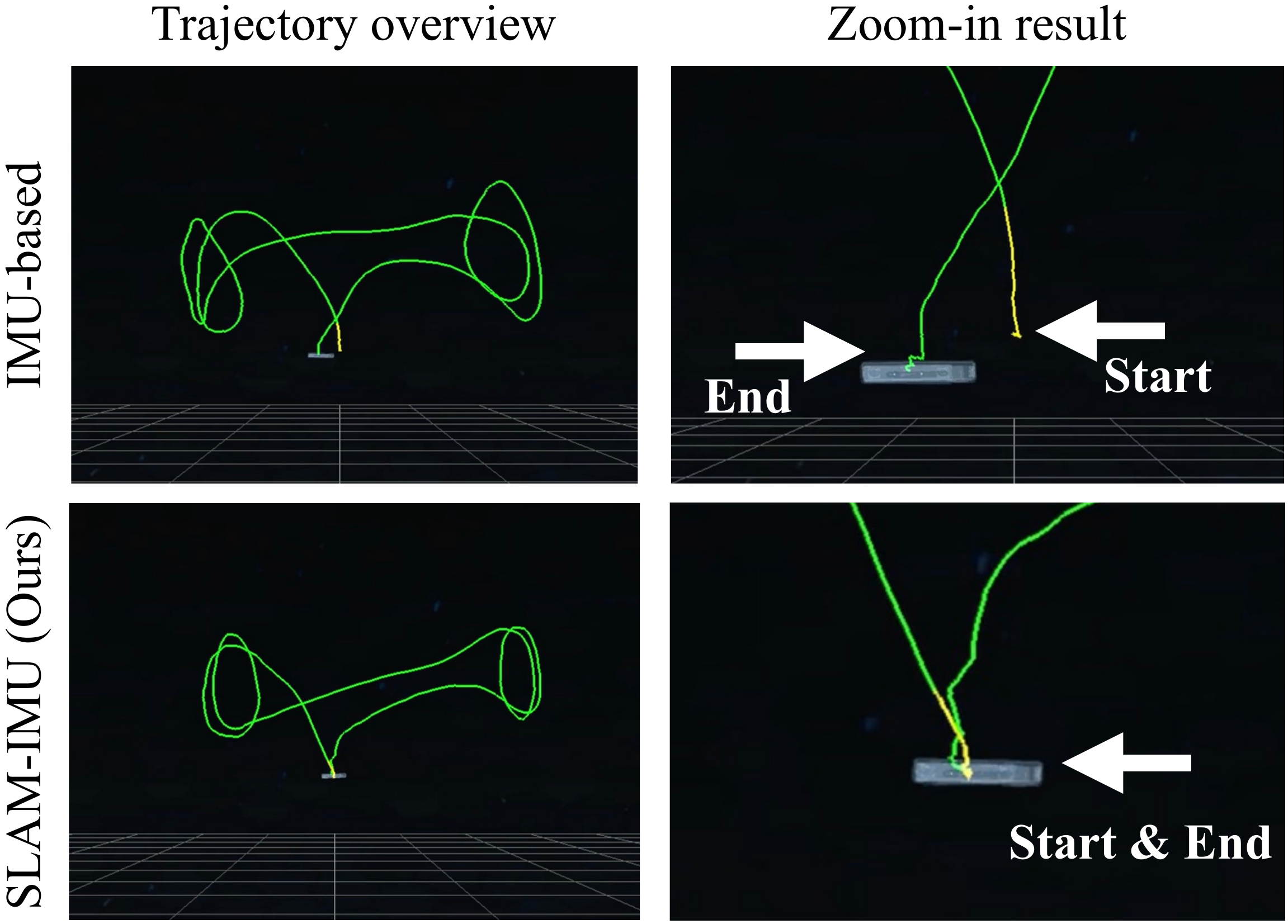}
\caption{\textbf{Compare with IMU-based mocap system.} We disable the SLAM mapping and pose-correction features of the T265 tracking camera, forcing it to rely on IMU information to track the pose. The human operator held the camera, started from a fixed location, moved it along a predefined trajectory, and then returned to the starting position. IMU-based method (first row) fails to match the endpoint with the start point, which indicates that there is pose drift during tracking. Our SLAM-IMU method (second row) doesn't drift and captures smooth trajectory during the tracking.}
\label{fig:exp_imu}
\end{figure}

\begin{table}[t]
\centering
\resizebox{\linewidth}{!}{
\footnotesize
\setlength{\tabcolsep}{4.0mm}{
\begin{tabular}{l|cc}
\toprule
\multicolumn{1}{c|}{Drifting error (cm)} & \multicolumn{1}{c}{Trajectory 1} & \multicolumn{1}{c}{Trajectory 2} \\
\midrule
IMU-based          & $8.0 \pm 3.1$ & $11.3 \pm 4.7$ \\
SLAM-IMU (Ours)         & $0.4 \pm 0.2$ & $0.8 \pm 0.3$ \\
\bottomrule    
\end{tabular}
}}
\caption{Drifting error of different tracking methods.}
\label{tab:drifting_error}
\end{table}

\subsection{Tracking accuracy} 
Figure \mbox{\ref{fig:exp_imu}} and Table \mbox{\ref{tab:drifting_error}} present qualitative and quantitative results, respectively. We observe that the IMU-based method suffers from pose drifting during tracking, while our SLAM-IMU approach more accurately tracks hand poses, with an average error of $0.8$ cm compared to the $11.3$ cm error of the IMU-based method.

\end{document}

%% file: figs/pull.tex
\twocolumn[{%
    \renewcommand\twocolumn[1][]{#1}%
        \maketitle
        \vspace{-5mm}
	\begin{center}

\vspace{-0cm}
\includegraphics[width=\textwidth]{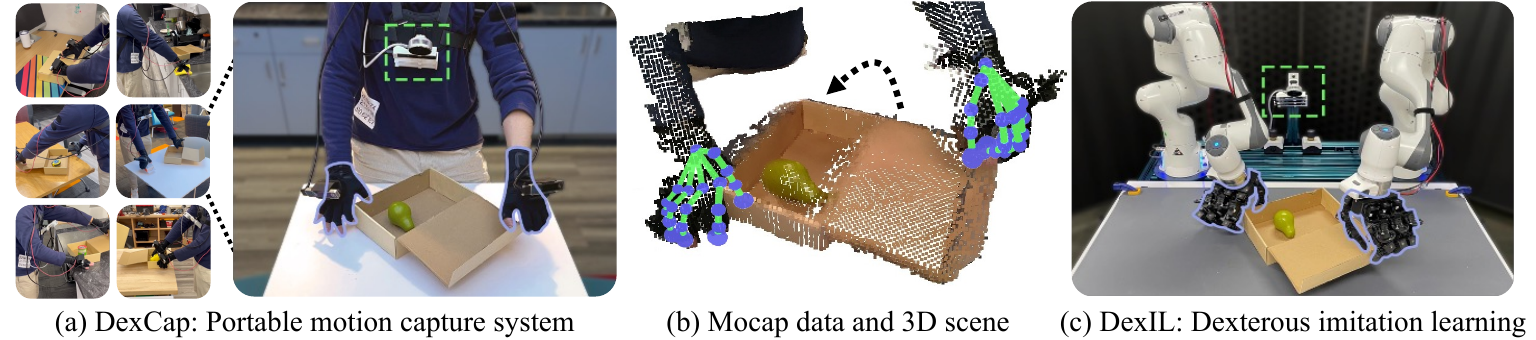}
    \captionof{figure}{
    \textbf{\sysName} facilitates the in-the-wild collection of high-quality human hand motion capture data and 3D observations. Leveraging this data, \textbf{\algoName} adapts it to the robot embodiment and trains control policy to perform the same task.}
    \label{fig:pull}
    \end{center}
}]